\documentclass[conference]{IEEEtran}

\newtheorem{theorem}{Theorem}[section]

\newtheorem{assumption}[theorem]{Assumption}
\newtheorem{problem}{Problem}
\newtheorem{definition}[theorem]{Definition}
\newtheorem{rem}[theorem]{Remark}

\usepackage{graphicx} 
\usepackage{epsfig} 
\usepackage{amsmath}
\usepackage{amssymb}
\usepackage{epstopdf}
\usepackage{cite}
\usepackage[noend,ruled,linesnumbered]{algorithm2e}
\SetKwComment{Comment}{$\triangleright$\ } {}
\usepackage{multirow}
\usepackage{rotating}
\usepackage{subfigure} 
\usepackage{color} 
\usepackage{mysymbol}
\usepackage[dvipsnames]{xcolor}
\usepackage[hyphens]{url}

\usepackage[breaklinks=true, colorlinks, bookmarks=true, citecolor=Black, urlcolor=Violet,linkcolor=Black]{hyperref}

\newcommand{\set}[1]{\left\{#1\right\}}

\usepackage{comment}
\usepackage[colorinlistoftodos,prependcaption,textwidth=1.5cm,textsize=tiny]{todonotes}

\IEEEoverridecommandlockouts                

\begin{document}
\title{\LARGE \bf Accelerated Reinforcement Learning \\ for Temporal Logic Control Objectives}
\author{ Yiannis Kantaros\thanks{The author is with the Department of Electrical and Systems
Engineering, Washington University in St. Louis, St. Louis, MO, 63130, USA. $\left\{\text{ioannisk}\right\}$@wustl.edu.  
}}

\maketitle 

\begin{abstract}
%
This paper addresses the problem of learning control policies for mobile robots, modeled as unknown Markov Decision Processes (MDPs), that are tasked with temporal logic missions, such as sequencing, coverage, or surveillance. The MDP captures uncertainty in the workspace structure and the outcomes of control decisions. The control objective is to synthesize a control policy that maximizes the probability of accomplishing a high-level task, specified as a Linear Temporal Logic (LTL) formula. To address this problem, we propose a novel accelerated model-based reinforcement learning (RL) algorithm for LTL control objectives that is capable of learning control policies significantly faster than related approaches. Its sample-efficiency relies on biasing exploration towards directions that may contribute to task satisfaction. This is accomplished by leveraging an automaton representation of the LTL task as well as a continuously learned MDP model. Finally, we provide comparative experiments that demonstrate the sample efficiency of the proposed method against recent RL methods for LTL objectives. 
\end{abstract}

%


   
\section{Introduction} \label{sec:Intro}
\vspace{-0.1cm}

Reinforcement learning (RL) has recently emerged as a prominent tool to synthesize optimal control policies for stochastic systems, modeled as Markov Decision Processes (MDPs), with complex control objectives captured by formal languages, such as Linear Temporal Logic (LTL); see e.g., \cite{hahn,gao2019reduced,bouton2019reinforcement,hasanbeig2019reinforcement,bozkurt2020control,cai2021reinforcement,lavaei2020formal,wang2020continuous,jothimurugan2021compositional} and the references therein. Common in the majority of these works is that they explore \textit{randomly} a product state space that grows exponentially as the size of the MDP and/or the complexity of the assigned temporal logic task increase. This results in sample inefficiency and slow training/learning process. This issue becomes more pronounced by the sparse rewards that these methods rely on to synthesize control policies with probabilistic satisfaction guarantees \cite{bozkurt2020control}.

To accelerate the learning process, several reward engineering methods have been proposed that augment the reward signal. For instance, hierarchical RL and reward machines have been proposed recently that rely on introducing rewards for intermediate goals \cite{icarte2018using,wen2020efficiency,toro2022reward}. Such methods often require a user to \textit{manually} decompose the global task into sub-tasks. Then additional rewards are assigned to these intermediate sub-tasks. Nevertheless, this may result in sub-optimal control policies with respect to the original task  \cite{zhai2022computational} while their efficiency highly depends on the task decomposition (i.e., the density of the rewards) \cite{cai2022overcoming}. 
We note that augmenting the reward signal for temporal logic tasks may compromise the probabilistic correctness of the synthesized controllers \cite{bozkurt2020control}.
Several methods that do not require modification of the reward signal, such as Boltzmann/softmax \cite{sutton1990integrated,kaelbling1996reinforcement,cesa2017boltzmann} and upper confidence bound (UCB) \cite{auer2002finite,chen2017ucb} have also been proposed to enhance sample efficiency; however, to the best of our knowledge, these techniques have not been extended and applied to temporal logic learning tasks.
%
A recent survey can be found in \cite{amin2021survey}. 

In this paper, we propose an accelerated model-based and value-based RL method that can quickly learn control policies for stochastic systems, modeled as unknown MDPs, with LTL control objectives. The unknown MDP has discrete state and action space and it models uncertainty in the workspace and in the outcome of control decisions. The sample-efficiency of the proposed algorithm relies on a novel \textit{logic-based exploration} strategy. 
%
%
%
We note that the proposed exploration strategy does not require knowledge or modification of the reward structure. Additionally, sample-efficiency of the proposed algorithm can be further improved by facilitating exploitation, as well, by e.g., introducing intermediate rewards  \cite{toro2022reward,cai2022overcoming}; nevertheless, this is out of the scope of this paper. Finally, we provide comparative experiments demonstrating that the proposed learning algorithm outperforms, in terms of sample-efficiency, RL methods that employ random (e.g., \cite{gao2019reduced,hasanbeig2019reinforcement, bozkurt2020control}), Boltzmann, and UCB exploration.

\textbf{Contribution:} \textit{First}, we propose a model-based RL algorithm that can quickly learn control policies for \textit{unknown} MDPs and LTL control objectives. \textit{Second}, we demonstrate how the automaton representation of any LTL task can be leveraged to enhance sample-efficiency. 
\textit{Third}, we provide comparative experiments demonstrating the sample efficiency of the proposed method against related RL methods for LTL objectives.

\vspace{-0.1cm}
\section{Problem Definition} \label{sec:PF}
\vspace{-0.1cm}
Consider a robot that resides in a partitioned environment with a finite number of states. To capture uncertainty in the robot motion and the workspace, we model the interaction of the robot with the environment as a Markov Decision Process (MDP) of unknown structure, which is defined as follows.

\begin{definition}[MDP]\label{def:labelMDP}
A Markov Decision Process (MDP) is a tuple $\mathfrak{M} = (\ccalX, x_0,  \ccalA, P, \mathcal{AP})$, where $\ccalX$ is a finite set of states; $x_0\in\ccalX$ is the initial state; $\ccalA$ is a finite set of actions. With slight abuse of notation $\ccalA(x)$ denotes the available actions at state $x\in\ccalX$; 
$P:\ccalX\times\ccalA\times\ccalX\rightarrow[0,1]$ is the transition probability function so that $P(x,a,x')$ is the transition probability from state $x\in\ccalX$ to state $x'\in\ccalX$ via control
action $a\in\ccalA$ and $\sum_{x'\in\ccalX}P(x,a,x')=1$, for all $a\in\ccalA(x)$; $\mathcal{AP}$ is a set of atomic propositions; $L : \ccalX \rightarrow 2^{\mathcal{AP}}$ is the labeling function that returns the atomic propositions that are satisfied at a state $x \in \ccalX$. 
\end{definition}

Hereafter, we make the following two assumptions about the MDP $\mathfrak{M}$:
\vspace{-0.1cm}
\begin{assumption}[Fully Observable MDP]
We assume that the MDP $\mathfrak{M}$ is fully observable, i.e., at any time/stage $t$ the current state, denoted by $x_t$, and the observations in state $x_t$, denoted by $\ell_t=L(x_t)\in 2^{\mathcal{AP}}$, are known. 
\end{assumption}
\vspace{-0.05cm}
\begin{assumption}[Labeling Function]\label{as:label}
We assume that the labeling function  $L : \ccalX \rightarrow 2^{\mathcal{AP}}$ is known, i.e., the robot knows which atomic propositions are satisfied at each MDP state. 
\end{assumption}

At any stage $T\geq0$ we define: (i) the robot's past path as $X_T=x_0x_1\dots x_T$; (ii) the past sequence of observed labels as $L_T=\ell_0\ell_1\dots \ell_T$, where $\ell_t\in 2^\mathcal{AP}$; (iii) and the past sequence of control actions $\ccalA_T=a_0a_1\dots a_{T-1}$, where $a_t\in\ccalA(x_t)$. These three sequences can be composed into a complete past run, defined as $R_T=x_0\ell_0 a_0 x_1\ell_1 a_1 \dots x_T\ell_T$. We denote by $\ccalX_T$, $\ccalL_T$, and $\ccalR_T$ the set of all possible sequences $X_T$, $L_T$ and $R_T$, respectively, starting from the initial MDP state $x_0$. 


The robot is responsible for accomplishing a task expressed as an LTL formula, such as sequencing, coverage, surveillance, data gathering or  connectivity tasks \cite{fainekos2005temporal, leahy2016persistent, guo2017distributed, kantaros2017distributed}. LTL is a formal language that comprises a set of atomic propositions $\mathcal{AP}$, the Boolean operators, i.e., conjunction $\wedge$ and negation $\neg$, and two temporal operators, next $\bigcirc$ and until $\cup$. LTL formulas over a set $\mathcal{AP}$ can be constructed based on the following grammar: $\phi::=\text{true}~|~\pi~|~\phi_1\wedge\phi_2~|~\neg\phi~|~\bigcirc\phi~|~\phi_1~\cup~\phi_2,$ where $\pi\in\mathcal{AP}$. The other Boolean and temporal operators, e.g., \textit{always} $\square$, have their standard syntax and meaning. An infinite \textit{word} $\sigma$ over the alphabet $2^{\mathcal{AP}}$ is defined as an infinite sequence  $\sigma=\pi_0\pi_1\pi_2\dots\in (2^{\mathcal{AP}})^{\omega}$, where $\omega$ denotes infinite repetition and $\pi_k\in2^{\mathcal{AP}}$, $\forall k\in\mathbb{N}$. The language $\left\{\sigma\in (2^{\mathcal{AP}})^{\omega}|\sigma\models\phi\right\}$ is defined as the set of words that satisfy the LTL formula $\phi$, where $\models\subseteq (2^{\mathcal{AP}})^{\omega}\times\phi$ is the satisfaction relation \cite{baier2008principles}. In what follows, we consider atomic propositions of the form $\pi^{x_i}$ that are true if the robot is at state $x_i\in\ccalX$ and false otherwise. 

Our goal is to compute a finite-memory policy  $\boldsymbol\xi$ for $\mathfrak{M}$ defined as $\boldsymbol\xi=\xi_0\xi_1\dots$, where 
$\xi_t:R_t\times\ccalA\rightarrow[0,1]$, and $R_t$ is the past run for all $t\geq 0$. Let $\boldsymbol\Xi$ be the set of all such policies. Given a control policy $\boldsymbol\xi\in\boldsymbol\Xi$, the probability measure $\mathbb{P}_\mathfrak{M}^{\boldsymbol\xi}$, defined on the smallest $\sigma$-algebra over $\ccalR_{\infty}$, 
is the unique measure defined as $\mathbb{P}_\mathfrak{M}^{\boldsymbol\xi}=\prod_{t=0}^T P(x_t,a_t,x_{t+1})\xi_t(x_t,a_t)$, where $\xi_t(x_t,a_t)$ denotes the probability of selecting the action $a_t$ at state $x_t$.
%
%
We then define the probability of $\mathfrak{M}$ satisfying $\phi$ under policy $\boldsymbol\xi$ as $\mathbb{P}_\mathfrak{M}^{\boldsymbol\xi}(\phi)=\mathbb{P}_\mathfrak{M}^{\boldsymbol\xi}(\ccalR_{\infty}:\ccalL_{\infty}\models\phi)$ \cite{baier2008principles}.
%
The problem we address in this paper is summarized as follows.
\begin{problem}\label{pr:pr1}
Given an MDP $\mathfrak{M}$ with unknown transition probabilities, unknown underlying graph structure, and a task specification captured by an LTL formula $\phi$, synthesize a finite memory control policy $\boldsymbol\xi^*$ that maximizes the probability of satisfying $\phi$, 
i.e., $\boldsymbol\xi^*=\argmax_{\boldsymbol\xi}\mathbb{P}_\mathfrak{M}^{\boldsymbol\xi}(\phi)$.
\end{problem}

\section{Accelerated Reinforcement Learning for Temporal Logic Control}
\vspace{-0.1cm}
\label{sec:planning}






\normalsize

To solve Problem \ref{pr:pr1}, we propose a new reinforcement learning (RL) algorithm that can quickly synthesize control policies that maximize the probability of satisfying LTL specifications. The proposed algorithm is summarized in Algorithm \ref{alg:RL-LTL} and it is described in detail in the following subsections. 


\subsection{From LTL formulas to DRA}\label{sec:DRA}
\vspace{-0.1cm}

Any LTL formula $\phi$, capturing a robot task, can be translated into a Deterministic Rabin Automaton (DRA) defined as follows [line \ref{rl:dra}, Alg. \ref{alg:RL-LTL}].
\begin{definition}[DRA \cite{baier2008principles}]\label{def:dra}
A DRA over $2^{\mathcal{AP}}$ is a tuple $\mathfrak{D} = (\ccalQ_D, q_D^0, \Sigma, \delta_D, \ccalF)$, where $\ccalQ_D$ is a finite set of states; $q_D^0 \subseteq Q_D$ is the initial state; $\Sigma=2^{\mathcal{AP}}$ is the input alphabet; $\delta_D: \ccalQ_D \times \Sigma_D \rightarrow {\ccalQ_D}$ is the transition function; and $\ccalF = \{  (\ccalG_1,\ccalB_1), \dots , (\ccalG_f,\ccalB_f) \}$ is a set of accepting pairs where $\ccalG_i, \ccalB_i \subseteq \ccalQ_D, \forall i \in\{1,\dots,f\}$. 
\end{definition}

An infinite run $\rho_{D}=q_D^0q_D^1\dots q_D^k\dots$ of $D$ over an infinite word $\sigma=\pi_0\pi_1\pi_2\dots$, where $\pi_k\in\Sigma=2^{\mathcal{AP}}$, $\forall k\in\mathbb{N}$, is an infinite sequence of DRA states $q_D^k$, $\forall k\in\mathbb{N}$, such that $\delta_D(q_D^k,\pi_k)=q_D^{k+1}$. An infinite run $\rho_D$ is called \textit{accepting} if there exists at least one pair $(\ccalG_i,\ccalB_i)$ such that $\texttt{Inf}(\rho_D)\cap\ccalG_i\neq\varnothing$  and $\texttt{Inf}(\rho_D)\cap\ccalB_i=\varnothing$, where $\texttt{Inf}(\rho_B)$ represents the set of states that appear in $\rho_D$ infinitely often.


\subsection{Distance Function over the DRA State Space}\label{sec:prune}
\vspace{-0.1cm}
In what follows, building upon \cite{kantaros2020stylus,kantaros2022perception}, we present a distance-like function over the DRA state-space that measures how `far' the robot is from accomplishing an assigned LTL task [line \ref{rl:dist}, Alg. \ref{alg:RL-LTL}]. In other words, this function returns how far any given DRA state is from the sets of accepting states $\ccalG_i$. To define this function, first, we prune the DRA $\mathfrak{D}$ by removing all infeasible transitions, i.e., transitions that cannot be enabled. To define infeasible transitions, we first define feasible symbols as follows. 
%

\begin{definition}[Feasible symbols $\sigma\in\Sigma$]\label{def:feas}
A symbol $\sigma\in\Sigma$ is \textit{feasible} if and only if $\sigma\not\models b^{\text{inf}}$, where $b^{\text{inf}}$ is a Boolean formula defined as  $b^{\text{inf}}=\vee_{\forall x_i\in\ccalX}( \vee_{\forall x_e\in\ccalX\setminus\{x_i\}}(\pi^{x_i}\wedge\pi^{x_e}))$
where $b^{\text{inf}}$ requires the robot to be present in more than one MDP state simultaneously. All feasible symbols $\sigma$ are collected in a set denoted by $\Sigma_{\text{feas}}\subseteq\Sigma$.
\end{definition}

Then, we prune the DRA by removing infeasible DRA transitions defined as follows:
\begin{definition}[Feasible \& Infeasible DRA transitions]
Assume that there exist $q_D, q_D'\in\ccalQ_D$ and $\sigma\in\Sigma$ such that $\delta_D(q_D,\sigma)=q_D'$. The DRA transition from $q_D$ to $q_D'$ is feasible if there exists at least one feasible symbol $\sigma\in\Sigma_{\text{feas}}$ such that $\delta_D(q_D,\sigma)=q_D'$; otherwise, it is infeasible.
\end{definition}


Next,  we define a function $d:\ccalQ_D\times\ccalQ_D\rightarrow \mathbb{N}$ that returns the minimum number of \textit{feasible} DRA transitions that are required to reach a state $q_D'\in\ccalQ_D$ starting from a state $q_D\in\ccalQ_D$. 
Particularly, we define the function $d: \ccalQ_D \times \ccalQ_D \rightarrow \mathbb{N}$ as follows
\begin{equation}\label{eq:dist1}
d(q_D,q_D')=\left\{
                \begin{array}{ll}
                  |SP_{q_D,q_D'}|, \mbox{if $SP_{q_D,q_D'}$ exists,}\\
                  \infty, ~~~~~~~~~\mbox{otherwise},
                \end{array}
              \right.
\end{equation}
where $SP_{q_D,q_D'}$ denotes the shortest path (in terms of hops) in the  pruned DRA from $q_D$ to $q_D'$ and $|SP_{q_D,q_D'}|$ stands for its cost (number of hops). Note that if $d(q_D^0,q_D)= \infty$, for all $q_D\in\ccalG_i$ and for all $i\in\{1,\dots,n\}$, then the LTL formula can not be satisfied. The reason is that in the pruning process, only the DRA transitions that are impossible to enable are removed (i.e., the ones that require the robot to be physically present at more than one MDP state, simultaneously). Next, using \eqref{eq:dist1}, we define the following distance function:

\begin{equation}\label{eq:dist2G}
    d_F(q_D,\ccalF)=\min_{q_D^G\in\cup_{i\in\{1,\dots,F\}}\ccalG_i} d(q_D,q_D^G)
\end{equation}

In words, \eqref{eq:dist2G} measures the distance of any DRA state $q_D$ to the set of accepting pairs. This distance is equal to the distance (as per \eqref{eq:dist1}) to closest  DRA state $q_D^G$ that belongs to $\cup_{i\in\{1,\dots,F\}}\ccalG_i$, i.e., to the union of accepting DRA states. 

\subsection{Product MDP}\label{sec:PMDP} 
\vspace{-0.1cm}
Given the MDP $\mathfrak{M}$ and the (non-pruned) DRA  $\mathfrak{D}$, we define the product MDP (PMDP) $\mathfrak{P}=\mathfrak{M} \times \mathfrak{D}$ as follows.

\begin{definition}[PMDP]\label{def:prodMDP}
Given an MDP $\mathfrak{M} \allowbreak = (\ccalX\allowbreak, x_0\allowbreak,  \ccalA\allowbreak, P\allowbreak, \mathcal{AP}\allowbreak)$ and a DRA $\mathfrak{D} = (\ccalQ_D, q_D^0, \Sigma, \ccalF, \delta_D)$, we define the product MDP (PMDP) $\mathfrak{P}=\mathfrak{M} \times \mathfrak{D}$ as  $\mathfrak{P}\allowbreak = (\mathcal{S}\allowbreak, {s}_0\allowbreak, \ccalA_\mathfrak{P}\allowbreak, P_{\mathfrak{P}}\allowbreak, \mathcal{F}_\mathfrak{P})$, where 
(i) $\mathcal{S} = \ccalX \times \ccalQ_D$ is the set of states, so that $s=(x,q_D)\in\ccalS$, $x\in\ccalX$, and $q_D\in\ccalQ_D$ ; 
(ii) ${s_0} = (x_0, q_D^0)$ is the initial state; 
(iii) $\mathcal{A}_\mathfrak{P}$ is the set of actions inherited from the MDP, so that $\mathcal{A}_\mathfrak{P} (s) = \ccalA(x)$, where $s=(x,q_D)$;  
(iv) $P_\mathfrak{P}:\ccalS\times\ccalA_\mathfrak{P}\times\ccalS:[0,1]$ is the transition probability function, so that $P_\mathfrak{P}(s,a_P,s')=P(x,a,x')$,
where $s=(x,q_D)\in\ccalS$, $s'=(x',q_D')\in\ccalS$, $a_P\in\ccalA(s)$ and $q_D'=\delta_D(q,L(x))$; 
(v) $\mathcal{F}_\mathfrak{P} = \{\mathcal{F}_i^\mathfrak{P}\}_{i=1}^{f}$ is the set of accepting states, where $\mathcal{F}_i^\mathfrak{P}$ is a set defined as $\mathcal{F}_i^\mathfrak{P} =  \ccalX \times  \mathcal{F}_i$ and $\mathcal{F}_i=(\ccalG_i,\ccalB_i)$.$\hfill\Box$
\end{definition}

Given any stationary and deterministic policy $\boldsymbol\mu:\ccalS\rightarrow \ccalA_\mathfrak{P}$ for $\mathfrak{P}$, we define an infinite run $\rho_{\mathfrak{P}}^{\boldsymbol\mu}$ of $\mathfrak{P}$ to be an infinite sequence of states of $\mathfrak{P}$, i.e., $\rho_{\mathfrak{P}}^{\boldsymbol\mu}=s_0s_1s_2\dots$, where $P_{\mathfrak{P}}(s_t,\boldsymbol\mu(s_t),s_{t+1})>0$. By definition of the accepting condition of the DRA $\mathfrak{D}$, an infinite run  $\rho_{\mathfrak{P}}^{\boldsymbol\mu}$ is accepting, i.e., $\boldsymbol\mu$ satisfies $\phi$ with a non-zero probability (denoted by $\boldsymbol\mu\models\phi$), if $\texttt{Inf}(\rho_{\mathfrak{P}}^{\boldsymbol\mu})\cap\ccalG^\mathfrak{P}_i\neq\emptyset$, and $\texttt{Inf}(\rho_{\mathfrak{P}}^{\boldsymbol\mu})\cap\ccalB^\mathfrak{P}_i=\emptyset$ $\forall i\in\set{1,\dots,f}$.

\subsection{Construction of the Reward Function}\label{sec:Reward}
\vspace{-0.1cm}
In what follows, we design a synchronous reward function based on the accepting condition of the PMDP so that maximization of the expected accumulated reward implies maximization of the satisfaction probability. Specifically, we adopt the reward function  $R:\ccalS\times\ccalA_{\mathfrak{P}}\times\ccalS$ defined in \cite{dorsa2014learning} constructed as follows:
\begin{equation}\label{eq:RewardQ}
R(s,a_\mathfrak{P},s')=\left\{
                \begin{array}{ll}
                  r_{\ccalG}, ~\mbox{if $s'\in\ccalG_i^{\mathfrak{P}}$,}\\
                  r_{\ccalB}, ~\mbox{if $s'\in\ccalB_i^{\mathfrak{P}}$,}\\
                  r_0, ~\mbox{otherwise}
                \end{array}
              \right.
\end{equation}
where $r_{\ccalG}>0$, for all $i\in\{1,\dots,f\}$, $r_{\ccalB}<r_0\leq0$. Using this reward function, the robot is motivated to satisfy the PMDP accepting condition, i.e., visit the states $\ccalG_j^{\mathfrak{P}}$ as often as possible and minimize the number of times it visits $\ccalB_i^{\mathfrak{P}}$ while following the shortest possible path. We note that any other reward function for LTL tasks can be employed (see e.g., \cite{gao2019reduced}) in the sense that the sample-efficiency of the proposed method does not rely on the reward structure; nevertheless, clearly, the designed rewards may affect the learned policy. 

\begin{algorithm}[t]
\footnotesize
\caption{Accelerated RL for LTL planning}
\LinesNumbered
\label{alg:RL-LTL}
\KwIn{ (i) initial MDP state $x_0$, (ii) LTL formula $\phi$}
Initialize: (i) $Q^{\boldsymbol\mu}(s,a)$ arbitrarily, (ii) $P(x,a,x')=0$, (iii) $c(x,a,x')=0$, (iv) $n(x,a)=0$, for all $x,x'\in\ccalX$ and $a\in\ccalA(x)$, and  (v) $n_{\mathfrak{P}}(s,a)=0$ for all $s'\in\ccalS$ and $a\in\ccalA_{\mathfrak{P}}(s)$\;
Convert $\phi$ to a DRA $\mathcal{D}$\;\label{rl:dra}
Construct distance function $d_F$ over the DRA as per \eqref{eq:dist2G}\; \label{rl:dist}
$\boldsymbol\mu=(\epsilon,\delta)-\text{greedy}(Q_{})$\;\label{rl:init1}
$\texttt{episode-number}=1$\; \label{rl:init3}
\While{$Q$ has not converged}{\label{rl:while1}
$\texttt{episode-number} = \texttt{episode-number} + 1$\; \label{rl:epi}
		${s}_\text{cur}={s}_0$\; \label{rl:initState}
		$\texttt{iteration}=1$\; \label{rl:initIterEpi}
        \While{$\texttt{iteration}<\tau$}{ \label{rl:while2}
        Pick action $a_{\text{cur}}$ as per \eqref{eq:policy}\;\label{rl:pickAction}
        Execute $a_{\text{cur}}$ and observe ${s}_{\text{after}}=(x_{\text{after}},q_{\text{after}})$, and $R({s}_\text{cur},a_{\text{cur}},{s}_{\text{after}})$\; \label{rl:exec}
        $n({x}_\text{cur},a_{\text{cur}}) = n({x}_\text{cur},a_{\text{cur}}) + 1$\;\label{rl:incrCounterStAct}
        $c({x}_\text{cur},a_{\text{cur}},{x}_\text{after}) = c({x}_\text{cur},a_{\text{cur}},{x}_\text{after}) + 1$\; \label{rl:incrCounterStActSt}
        $ \hat{P}(x_{\text{cur}},a_{\text{cur}},x_{\text{after}})=\frac{c(x_{\text{cur}},a_{\text{cur}},x_{\text{after}})}{n(x_{\text{cur}},a_{\text{cur}})}$ \;\label{rl:probEst}
        $n_{\mathfrak{P}}({s}_\text{cur},a_{\text{cur}},{s}_\text{after}) = n_{\mathfrak{P}}({s}_\text{cur},a_{\text{cur}},{s}_\text{after}) + 1$\; \label{rl:incrCounterTrans}
        $Q^{\boldsymbol\mu}({s}_\text{cur},a_{\text{cur}})= Q^{\boldsymbol\mu}({s}_\text{cur},a_{\text{cur}})+(1/n_P({s}_\text{cur},a_{\text{cur}}))[R({s}_\text{cur},a_{\text{cur}})-Q^{\boldsymbol\mu}({s}_\text{cur},a_{\text{cur}})+\gamma \max_{a'}Q^{\boldsymbol\mu}({s}_{\text{next}},a'))]$\;\label{rl:updQ}
        ${s}_\text{cur}={s}_{\text{next}}$\; \label{rl:resetSt}
        $\texttt{iteration}= \texttt{iteration} + 1$\; \label{rl:iter0}
    	Update $\epsilon, \delta_b, \delta_e$\;  \label{rl:iter}
    	}}
\end{algorithm}
\normalsize

\subsection{Accelerated Learning of Control Policies for LTL Tasks}\label{sec:accRL}
\vspace{-0.1cm}
In this section, we present the proposed accelerated model-based RL algorithm for LTL control synthesis [lines \ref{rl:init1}-\ref{rl:iter}, Alg. \ref{alg:RL-LTL}]. The output of the proposed algorithm is a \textit{stationary} and  \textit{deterministic} policy $\boldsymbol\mu^*$ 
for $\mathfrak{P}$. Projection of $\boldsymbol\mu^*$ onto the MDP $\mathfrak{M}$ yields the finite memory policy $\boldsymbol\xi^*$ (see Problem \eqref{pr:pr1}). The policy $\boldsymbol\mu^*$ is designed so that it maximizes the expected accumulated return, i.e., $\boldsymbol\mu^*(s)=\arg\max\limits_{\boldsymbol\mu \in \mathcal{D}}~ {U}^{\boldsymbol\mu}(s)$,
%
where $\mathcal{D}$ is the set of all stationary deterministic policies over $\mathcal{S}$, and 	
\begin{equation}\label{eq:utility}
{U}^{\boldsymbol\mu}(s)=\mathbb{E}^{\boldsymbol\mu} [\sum\limits_{n=0}^{\infty} \gamma^n~ R(s_n,\boldsymbol\mu(s_n),s_{n+1})|s=s_0].
\end{equation}
In \eqref{eq:utility}, $\mathbb{E}^{\boldsymbol\mu} [\cdot]$ denotes the expected value given that the product MDP follows the policy $\boldsymbol\mu$ \cite{puterman}, $0\leq\gamma< 1$ is the discount factor, and $s_0,...,s_n$ is the sequence of states generated by policy $\boldsymbol\mu$ up to time step $n$, initialized at $s_0$. Note that for finite state MDPs, the optimal policy $\boldsymbol\mu^*$, if it exists, is stationary and deterministic \cite{puterman}.




To construct $\boldsymbol\mu^*$,  we employ episodic  Q-learning, a model-based RL algorithm [lines \ref{rl:init1}-\ref{rl:iter}, Alg. \ref{alg:RL-LTL}] \cite{rlbook}. 
%
%
Similar to standard Q-learning, starting from an initial PMDP state, we define learning episodes over which the robot picks actions as per a stationary and stochastic control policy $\boldsymbol\mu:\ccalS\times \ccalA_{\mathfrak{P}}\rightarrow [0,1]$ that eventually converges to $\boldsymbol\mu^*$ [lines \ref{rl:init1}-\ref{rl:init3}, Alg. \ref{alg:RL-LTL}]. During each episode the robot estimates the MDP transition probabilities as well; the estimated transition probabilities are denoted by $\hat{P}(x_{\text{cur}},a_{\text{cur}},x_{\text{after}})$ [line \ref{rl:probEst}, Alg. \ref{alg:RL-LTL}]. Each episode terminates after a user-specified number of iterations $\tau$ or if the robot reaches a deadlock PMDP state, i.e., a state with no outgoing transitions [lines \ref{rl:epi}-\ref{rl:iter}, Alg. \ref{alg:RL-LTL}]. The RL algorithm terminates once an action value function $Q^{\boldsymbol\mu}(s,a)$ has converged. This action value function is defined as the expected return for taking
action $a$ when at state $s$ and then following policy $\boldsymbol\mu$ \cite{rlbook}, i.e., $Q^{\boldsymbol\mu}(s,a)=\mathbb{E}^{\boldsymbol\mu} [\sum\limits_{n=0}^{\infty} \gamma^n~ R(s_n,\boldsymbol\mu(s_n),s_{n+1})|s_0=s, a_0=a]$.
We have that  $U^{\boldsymbol\mu}(s)=\max_{a \in \mathcal{A}_\mathfrak{P}(s)}Q^{\boldsymbol\mu}(s,a)$ \cite{rlbook}. The action-value function $Q^{\boldsymbol\mu}(s,a)$ can be initialized arbitrarily. 

As a policy $\boldsymbol\mu$, we propose an extension of the $\epsilon$-greedy policy, called $(\epsilon,\delta)$-greedy policy, that selects an action $a$ at an PMDP state $s$ by using the learned action-value function $Q^{\boldsymbol\mu}(s,a)$ and the continuously learned robot dynamics captured by the estimated transition probabilities $\hat{P}(x,a,x')$. Formally, the $(\epsilon,\delta)$-greedy policy $\boldsymbol\mu$ is defined as
%
\begin{equation}\label{eq:policy}
\boldsymbol\mu(s) = \begin{cases}
1-\epsilon + \frac{\delta_e}{|\mathcal{A}_\mathfrak{P}(s)}| &\text{if~}  a=a^*~\text{and~} a\neq a_b, \\
1-\epsilon + \frac{\delta_e}{|\mathcal{A}_\mathfrak{P}(s)}| +\delta_b &\text{if~}  a=a^*~\text{and~} a=a_b, \\
\delta_e/|\mathcal{A}_\mathfrak{P}(s)|  &\text{if}~a \in \mathcal{A}_\mathfrak{P}(s),\\
\delta_b &\text{if~} a=a_b
\end{cases}
\end{equation}
where $\epsilon,\delta_b,\delta_m\in[0,1]$ and $\epsilon = \delta_b+\delta_m$. In words, according to this policy, (i) with probability $1-\epsilon$, the \textit{greedy} action $a^*=\argmax_{a\in\mathcal{A}_\mathfrak{P}}Q^{\boldsymbol\mu}(s,a)$ is taken (as in the standard $\epsilon$-greedy policy); and (ii) an exploratory action is selected with probability $\epsilon=\delta_b+\delta_e$. The exploration strategy is defined as follows: (ii.1) with probability $\delta_e$ a random action $a$ is selected (\textit{random} exploration); and (ii.2) with probability $\delta_b$ the action, denoted by $a_b$, that is most likely to drive the robot towards an accepting product state in $\ccalG_i^{\mathfrak{P}}$ is taken (\textit{biased} exploration). The action $a_b$ will be defined formally in Section \ref{sec:biasedExpl}.
The parameter $\epsilon$ is selected  so that eventually all actions have been applied infinitely often at all states while $\epsilon$ converges to $0$ \cite{rlbook}. This way, we have that $\boldsymbol\mu$ asymptotically converges to the optimal greedy policy $\boldsymbol\mu^*=\argmax_{a\in\mathcal{A}_\mathfrak{P}} Q^*(s,a)$,
where $Q^*$ is the optimal action value function. 
%
%

\subsection{Biased Exploration for Accelerated Learning}\label{sec:biasedExpl}
\vspace{-0.1cm}
In what follows, we describe in detail the biased exploration component of the control policy \eqref{eq:policy}.
Particularly, let $s_{\text{cur}}=(x_{\text{cur}},q_{\text{cur}})$ denote the current  PMDP state at the current learning episode and iteration of Algorithm \ref{alg:RL-LTL}. To design the action $a_b$, for biased exploration, we first introduce the following definitions. 

Let $\ccalQ_{\text{goal}}(q_{\text{cur}})\subset\ccalQ$ be a set that collects all DRA states that are one-hop reachable from $q_{\text{cur}}$ in the pruned DRA and they are closer to the accepting DRA states than $q_{\text{cur}}$ is, as per \eqref{eq:dist2G}. In math, $\ccalQ_{\text{goal}}(q_{\text{cur}})$ is defined as follows: $\ccalQ_{\text{goal}}(q_{\text{cur}})=\{q'\in\ccalQ~|~(\exists\sigma\in\Sigma_{\text{feas}}~\text{such that}~\delta_D(q_{\text{cur}},\sigma)=q') \wedge (d_F(q',\ccalF)=d_F(q_{\text{cur}},\ccalF)-1)\}.$
%
%
%
%
Also, let $\ccalX_{\text{goal}}(q_{\text{cur}})\subseteq\ccalX$ be a set of MDP states, denoted by $x_{\text{goal}}$, that if the robot eventually reaches, then transition from $s_{\text{cur}}$ to a product state $s_{\text{goal}}=[x_{\text{goal}},q_{\text{goal}}]$ will occur, where  $q_{\text{goal}}\in\ccalQ_{\text{goal}}(q_{\text{cur}})$. Formally, $\ccalX_{\text{goal}}(s_{\text{cur}})$ is defined as follows:
\begin{equation}
\ccalX_{\text{goal}}(q_{\text{cur}}) = \{x\in\ccalX~|~\delta_D(q_{\text{cur}},L(x))=q_{\text{goal}}\in\ccalQ_{\text{goal}}(q_{\text{cur}})\}
\end{equation}

Next, for each $x_{\text{goal}}\in\ccalX_{\text{goal}}(q_{\text{cur}})$, we compute all MDP states, collected in a set denoted $\ccalX_{\text{closer}}(x_{\text{cur}})\subseteq\ccalX_{\text{goal}}(q_{\text{cur}})$, that are one hop reachable from $x_{\text{cur}}$ and they are closer to $\ccalX_{\text{goal}}(x_{\text{cur}})$ than $x_{\text{cur}}$ is. 
Specifically, we construct $\ccalX_{\text{closer}}(x_{\text{cur}})$ as follows. 
First, we compute the reachable set $\ccalR(x_{\text{cur}})\subseteq\ccalX$ that collects all MDP states that can be reached within one hop from $x_{\text{cur}}$, i.e., $ \ccalR(x_{\text{cur}}) = \{x\in\ccalX~|~\exists a\in\ccalA(x)~ \text{such that}~\hat{P}(x_{\text{cur}},a,x)>0\}$. Note that this reachable set is a subset of the actual reachable set since the former uses the estimated, and not the actual, transition probabilities.
%
Next, we view the continuously learned MDP as a weighted directed graph $\ccalG=(\ccalV,\ccalE, w)$ where the set $\ccalV$ is the set of MDP states, $\ccalE$ is the set of edges, and $w:\ccalE\rightarrow \mathbb{R}_{+}$ is function assigning weights to each edge. Specifically, an edge from the node (MDP state) $x$ to $x'$ exists if there exists at least one action $a\in\ccalA(x)$ such that $\hat{P}(x,a,x')>0$. Hereafter, we assigned a weight equal to $1$ to each edge; see also Remark \ref{rem:weight}. Also, we denote the cost of the shortest path from $x$ to $x'$ by $J_{x,x'}$.  Next, we define the cost of the shortest path connecting state $x$ to the set $\ccalX_{\text{goal}}$ as follows: $J_{x,\ccalX_{\text{goal}}}=\min_{x'\in\ccalX_{\text{goal}}} J_{x,x'}.$
Then, we define the set $\ccalX_{\text{closer}}$ as follows:
\begin{equation}
    \ccalX_{\text{closer}}(x_{\text{cur}})=\{x\in\ccalR(x_{\text{cur}})~|~J_{x,\ccalX_{\text{goal}}}=J_{x_{\text{cur}},\ccalX_{\text{goal}}}-1)\}.
\end{equation}

Once $ \ccalX_{\text{closer}}$ is constructed, the biased action $a_b$ is defined as follows:
\begin{equation}\label{eq:a_b}
 [a_b, x_b] = \argmax_{a\in \ccalA(x_{\text{cur}}), x\in\ccalX_{\text{closer}}(x_{\text{cur}})}\ \hat{P}(x_{\text{cur}},a,x).
\end{equation} 
In words, as per \eqref{eq:a_b}, among all transition probabilities $\hat{P}(x_{\text{cur}},a,x)$ associated with actions $a\in \ccalA(x_{\text{cur}})$ and states $x\in\ccalX_{\text{closer}}(x_{\text{cur}})$, the largest one is $\hat{P}(x_{\text{cur}},a_b,x_b)$. The MDP state $x_b$ is the state towards which $a_b$ is biased to.
%

\begin{rem}[Selecting exploration parameters $\delta_b$ and $\delta_e$]\label{rem:deltas}
The biased action $a_b$ is selected so that it drives the robots closer to $\ccalX_{\text{goal}}$ which requires the MDP model. Since the latter is unknown, the actual estimated transition probabilities are used in \eqref{eq:a_b} instead. As a result, the selected biased action $a_b$ may not be the one that would have been computed in \eqref{eq:a_b} if the transition probabilities were known. Thus, we select initially $\delta_e>\delta_b$ while $\delta_e$ converges to $0$ faster than $\delta_b$. Intuitively, this allows to initially perform random exploration to learn an accurate enough MDP model which is then exploited to bias exploration towards directions that will drive the system closer to the DRA accepting states. 
\end{rem}

\begin{rem}[Computing Shortest Path]\label{rem:shortestPath}
It is possible that the shortest path from $x_{\text{cur}}$ to $x_{\text{goal}}\in\ccalX_{\text{goal}}(q_{\text{cur}})$ goes through states/nodes $x$ that if visited, a transition to a new state $q\neq q_{\text{cur}}$ that does not belong to $\ccalQ_{\text{goal}}(q_{\text{cur}})$ may be enabled. Therefore, to compute the shortest paths for the construction of $\ccalX_{\text{closer}}(x_{\text{cur}})$, we treat all such nodes $x$ as `obstacles' that should not be crossed. These states are collected in the set $\ccalX_{\text{avoid}}$ defined as $\ccalX_{\text{avoid}}=\{x\in\ccalX~|~\delta(q_{\text{cur}},L(x))=q_D\notin\ccalQ_{\text{goal}}\}$.
\end{rem}

\begin{rem}[Weights]\label{rem:weight}
To design the biased action $a_b$, the MDP is viewed as weighted graph where a weight $w=1$ is assigned to all edges. 
Alternative weight assignments can be used as well. For instance, the assigned weights can be equal to the reciprocal of the estimated transition probabilities. In this case, the shortest path between two MDP states models the path with least uncertainty that connects these two states. We emphasize that the weight definition affects which states are included in $\ccalX_{\text{closer}}$. 
\end{rem}

\section{Numerical Experiments} \label{sec:Sim}
\vspace{-0.1cm}
In this section we present three case studies, implemented on MATLAB R2016a on a computer with an Intel Xeon CPU at 2.93\,GHz and 4\,GB RAM. In all experiments, the environment is represented as a $10\times 10$ discrete grid world, i.e., the interaction between the robot and the environment is modeled as an MDP $\mathfrak{M}$ with $|\ccalX|=100$ states. At each MDP state the robot can apply five actions: $\ccalA=\{\texttt{left}, \texttt{right}, \texttt{up}, \texttt{down}, \texttt{idle}\}$. The transition probabilities are designed so that  the probability of reaching the intended state is $0.7$ while the probability of reaching the remaining neighboring MDP states (including the current one) is $0.3/4=0.0750$. 

\begin{figure*}
    \centering
     \subfigure[]{
    \includegraphics[width=0.32\linewidth]{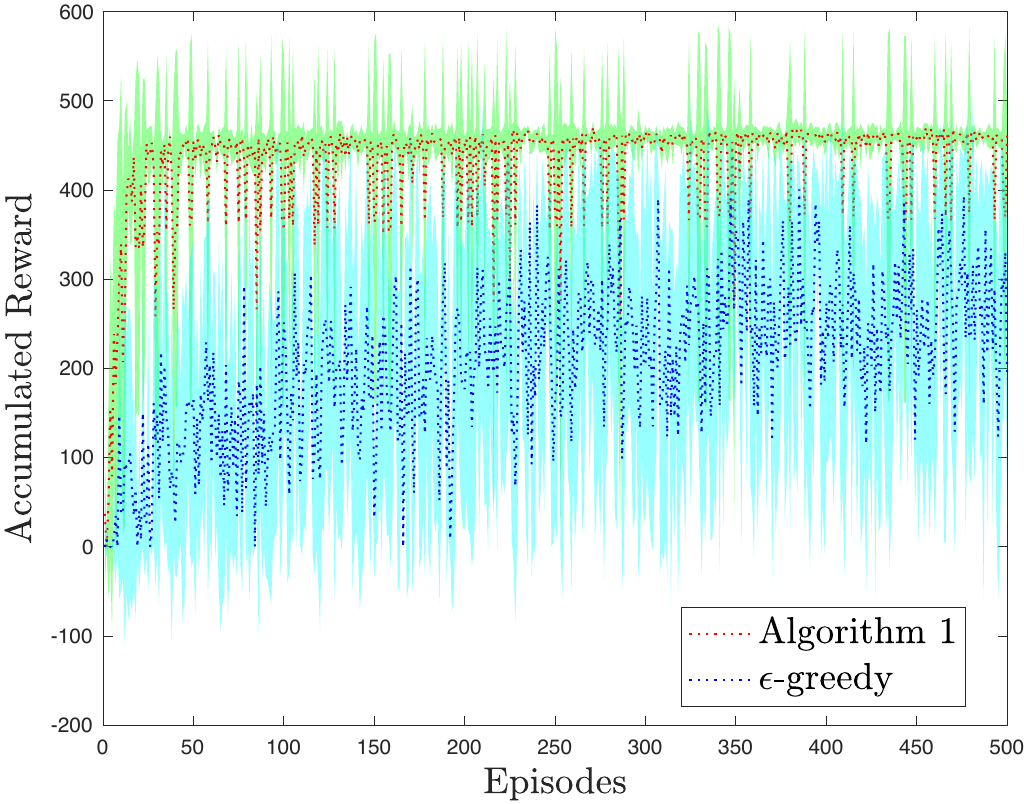}}
     \subfigure[]{
     \includegraphics[width=0.32\linewidth]{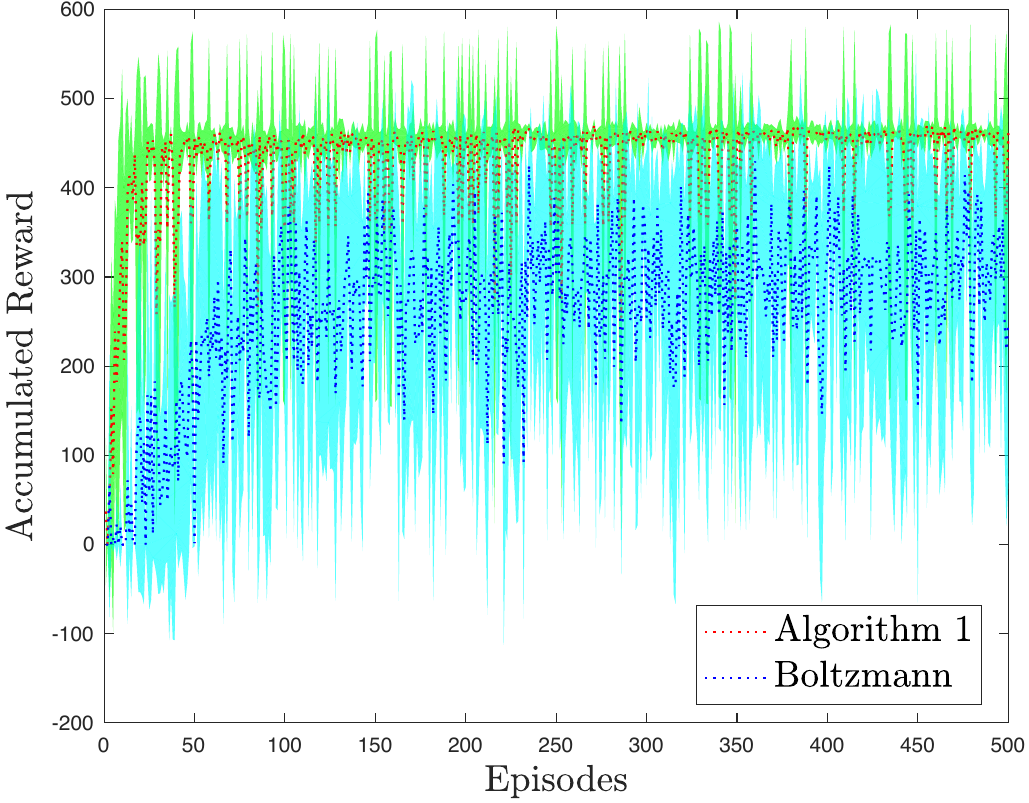}}
    \subfigure[]{
     \includegraphics[width=0.32\linewidth]{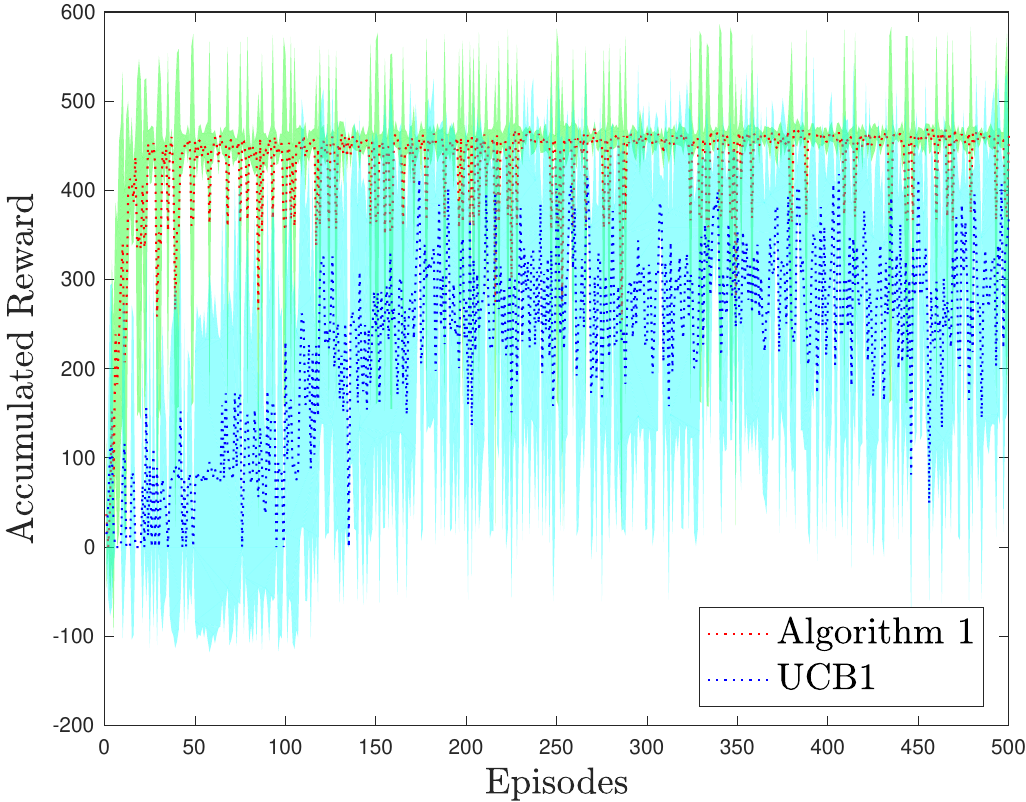}}
    \caption{Reach-Avoid Task I: Comparison of average accumulated reward over five runs of Algorithm \ref{alg:RL-LTL}, when it is applied with the proposed $(\epsilon,\delta)$-greedy policy, $\epsilon$-greedy policy, Boltzmann policy, and UCB1 policy. The red and blue curves illustrate the average accumulated rewards while the green and cyan regions denote the variance of the accumulated rewards.} \vspace{-5mm}
    \label{fig:task1}
\end{figure*}

\begin{figure}
    \centering
     \subfigure[Reach-Avoid Task II]{
     \label{fig:CompReachMoreObs}
    \includegraphics[width=0.48\linewidth] {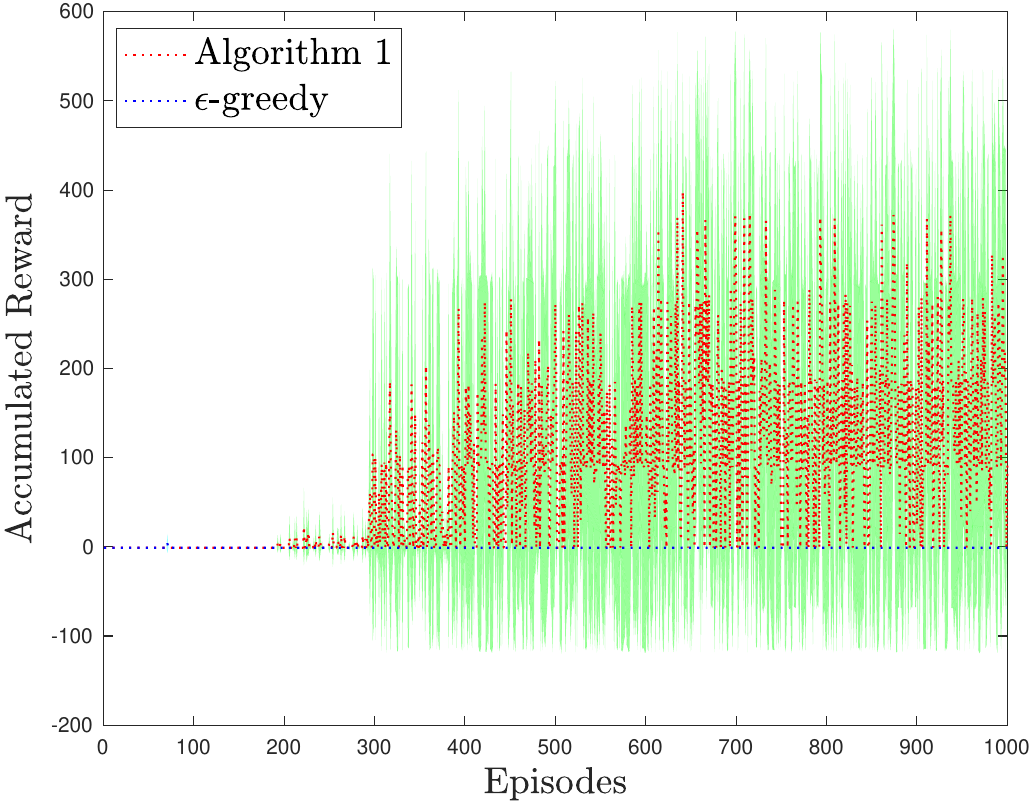}}
         \subfigure[Surveillance Task]{
           \label{fig:task3}
    \includegraphics[width=0.48\linewidth]{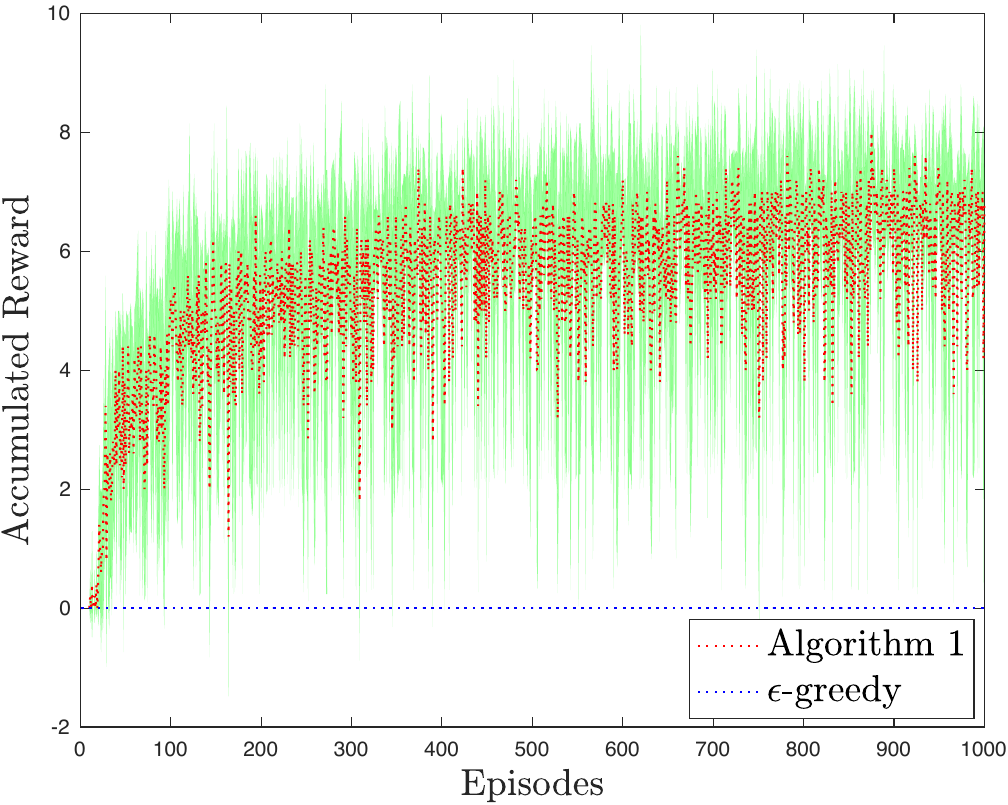}}
    \caption{Figures \ref{fig:CompReachMoreObs}-\ref{fig:task3} compare the average accumulated reward over five runs of Algorithm \ref{alg:RL-LTL}, for the reach-avoid task II and the surveillance task, when it is applied with the proposed $(\epsilon,\delta)$-greedy policy, $\epsilon$-greedy policy, Boltzmann policy, and UCB1 policy. The red and blue curves illustrate the average accumulated rewards. The green regions denote the variance of the accumulated rewards. All policies, besides the $(\epsilon,\delta)$-greedy policy, collect zero rewards within the first $1,000$ episodes.}\vspace{-5mm}
    \label{fig:twoTasks}
\end{figure}


In the following case studies we demonstrate the performance of Algorithm \ref{alg:RL-LTL} when it is equipped with the proposed $(\epsilon,\delta)$-greedy policy \eqref{eq:policy}, the $\epsilon$-greedy policy, the Boltzman policy, and the UCB1 policy. To make the comparison between the $(\epsilon,\delta)$- and the $\epsilon$-greedy policy fair, we select the same $\epsilon$ for both.   
The Boltzmann control policy is defined as follows: $\boldsymbol\mu_B(s) = \frac{e^{Q^{\boldsymbol\mu_B}(s,a)/T}}{\sum_{a'\in\ccalA_{\mathfrak{P}}}e^{Q^{\boldsymbol\mu_B}(s,a')/T}}$, 
%
where $T\geq 0$ is the temperature parameter used in the Boltzmann distribution \cite{kaelbling1996reinforcement,cesa2017boltzmann}. 
The UCB1 control policy is defined as:
$\boldsymbol\mu_U(s) = \argmax_{a\in\ccalA_{\mathfrak{P}}}\left[Q^{\boldsymbol\mu_U}(s,a) + C\times \sqrt{\frac{2\log(N(s))}{n(s,a)}}\right]$, where (i) $N(s)$ and $n(s,a)$ denote the number of times state $s$ has been visited and the number of times action $a$ has been selected at state $s$ and (ii) $C$ is an exploration parameter, i.e., the higher $C$ is, the more often exploration is applied \cite{auer2002finite,saito2014discounted}. 
%
Given any of these four policies $\boldsymbol\mu$, we run Algorithm \ref{alg:RL-LTL} for  $1000$ episodes and we compute \eqref{eq:utility}. Each episode runs for at most $\tau=500$ iterations. This process is repeated five times. Then, we report the average accumulated reward and its variance. 
In all case studies, we select $\gamma=0.99$ and $r_{\ccalG}=1$, $r_{\ccalB}=-10^{-4}$, and $r_o=0$ (see \eqref{eq:RewardQ}).  




\textbf{Reach-Avoid Task I:} First, we consider a simple reach-avoid task, where the robot has to reach the MDP state/region $x=100$ and stay there forever while always avoiding $x=46$ modeling an obstacle in the environment. This task can be captured by the following LTL formula: $\phi=\Diamond \square \pi^{100} \wedge \square \neg \pi^{46}$.
This LTL formula corresponds to a DRA with $4$ states and $1$ accepting pair. Thus, the product MDP consists of $100\times 4=400$ states.
%
%
To compare the performance of all four control policies, we compute the average discounted accumulated rewards, as discussed before, which is illustrated in Figure \ref{fig:task1}. 
Observe in this figure that the $(\epsilon,\delta)$-greedy policy outperforms all other policies. Also, notice that the UCB1 and the $\epsilon$-greedy policy attain similar performance while the Boltzman policy performs better than both. 
%
%

\textbf{Reach-Avoid Task II:} Second, we consider a more complex reach-avoid task with a larger number of obstacles. Specifically, the robot has to reach the MDP state/region $x=100$ and stay there forever while always avoiding $17$ obstacle cells. This task can be captured by a similar LTL formula as the one considered in the previous case study corresponding to a DRA with $4$ states and $1$ accepting pair. Thus, the product MDP consists of $100\times 4=400$ states.
%
%
The average discounted accumulated rewards over five runs of each RL algorithm is illustrated in Figure \ref{fig:CompReachMoreObs}. 
Observe in this figure that the $(\epsilon,\delta)$-greedy policy outperforms all other policies. In fact, notice that the UCB1, Boltzmann, and the $\epsilon$-greedy policy collect zero rewards within the first $1,000$ episodes. Compared to the previous case study, observe that the robot needs approximately $200$ more learning episodes to start collecting non-zero rewards. This is due to the higher number of obstacles in the environment. 

\textbf{Surveillance Task:} Third, we consider a surveillance/recurrence mission  captured by the following LTL formula: $ \phi=\square\Diamond \pi^{36} \wedge \square\Diamond \pi^{26} \wedge \square\Diamond \pi^{76} \wedge \square\Diamond \pi^{64} \wedge \square\Diamond \pi^{89} \wedge \square\Diamond \pi^{10} \wedge \square \neg \pi^{33}$.
In words, this task requires the robot to (i) patrol, i.e., to visit infinitely often and in any order the MDP states $36$, $26$, $76$, $64$, $89$, and $10$; (ii) and always avoid the state $33$ modeling an obstacle in the environment. The corresponding DRA has $14$ states and $1$ accepting pair resulting in a PMDP with $100\times 14=1,400$ states.
The comparative results are shown in Figure \ref{fig:task3}. Observe that the $(\epsilon,\delta)$-greedy policy performs significantly better than other approaches. In fact, after $1,000$ episodes the competitive approaches have collected zero rewards as opposed to the $(\epsilon,\delta)$-greedy policy that has started learning how to accumulate such sparse rewards very fast, i.e., only after few tens of episodes. We note that the considered task is more challenging than the previous ones as it requires the robot to visit a larger number of states; this increased complexity is also reflected in the size of the DRA state-space.

\section{Conclusion} \label{sec:Concl}
\vspace{-0.1cm}
In this paper, we proposed a new accelerated RL algorithm for LTL control objectives. Its sample efficiency relies on biasing exploration in the vicinity of task-related regions as supported by our comparative experiments. Our future work will focus on demonstrating sample-efficiency theoretically and enhancing scalability by using function approximations (e.g., neural networks). 
%



\bibliographystyle{IEEEtran}
\bibliography{YK_bib.bib}

\end{document}